\pdfoutput=1

\documentclass[11pt]{article}

\usepackage[preprint]{acl}

\usepackage{hyperref}
\usepackage{url}
\usepackage{times}
\usepackage{latexsym}
\usepackage{comment}
\usepackage{float}

\usepackage[utf8]{inputenc}
\usepackage{tabularx}
\usepackage{caption}
\usepackage{xurl}
\usepackage{booktabs}

\usepackage{amsmath}
\usepackage[T1]{fontenc}

\usepackage[utf8]{inputenc}

\usepackage{microtype}

\usepackage{inconsolata}

\usepackage{graphicx}

\title{The Rise of AI-Generated Content in Wikipedia}

\author{
    Creston Brooks \quad Samuel Eggert \quad Denis Peskoff \\
    Princeton University \\
    \texttt{\{cabrooks, sameggert, dp2896\}@princeton.edu}
}

\begin{document}
\maketitle
\begin{abstract}

The rise of AI-generated content in popular information sources raises significant concerns about accountability, accuracy, and bias amplification. Beyond directly impacting consumers, the widespread presence of this content poses questions for the long-term viability of training language models on vast internet sweeps. We use \texttt{GPTZero}, a proprietary AI detector, and \texttt{Binoculars}, an open-source alternative, to establish lower bounds on the presence of AI-generated content in recently created Wikipedia pages. Both detectors reveal a marked increase in AI-generated content in recent pages compared to those from before the release of GPT-3.5. With thresholds calibrated to achieve a 1\% false positive rate on pre-GPT-3.5 articles, detectors flag over 5\% of newly created English Wikipedia articles as AI-generated, with lower percentages for German, French, and Italian articles. Flagged Wikipedia articles are typically of lower quality and are often self-promotional or partial towards a specific viewpoint on controversial topics.

\end{abstract}

\section{AI-Generated Content}

As Large Language Models (LLMs) have become increasingly advanced and more accessible, the risks of convincingly generated text grow in tandem with the benefits.
 While benefits include easier communication through machine translation, increased productivity, and new pedagogical opportunities, risks include the increased scale of disinformation and misinformation \citep{goldstein2023generative}.
 Unchecked resampling of AI-generated data for training can even, in extreme cases, cripple model performance \citep{shumailov2024ai}. 
 Risks can be mitigated, however, to the extent that AI-generated data can be detected reliably at scale. 

%

\begin{figure}[t]
  \centering
  \includegraphics[width=\linewidth]{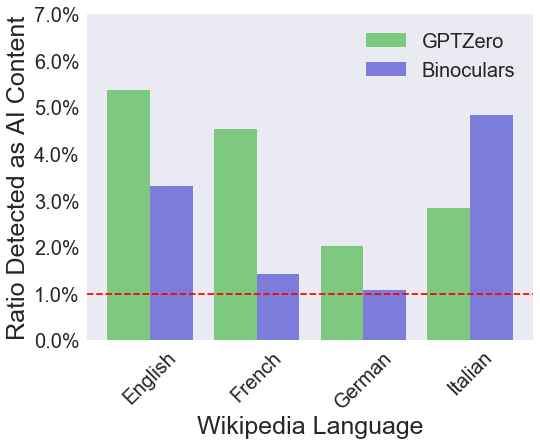} 
  \caption {Using two tools, \texttt{GPTZero} and \texttt{Binoculars}, we detect that as many as 5\% of 2,909 English Wikipedia articles created in August 2024 contain significant AI-generated content. The classification thresholds of both tools were calibrated to maintain a FPR of no more than 1\% on a pre-GPT-3.5 Wikipedia baseline, as indicated by the red line.}
\label{fig:wiki}
\end{figure}

With the rapid release of generative LLMs, AI detection has been developing in parallel \citep{tang2024science}. Individuals \citep{ferrara2024genai}, educators \citep{baidoo2023education,khalil2023will}, companies~\citep{jabeur2023artificial, adelani2020generating}, and governments~\citep{androutsopoulou2019transforming} seek reliable ways of validating that content has been generated by human authors rather than machines. 
Nonetheless, evaluating AI detectors across diverse contexts (e.g., length, domain, and level of integration with human writing) remains challenging ~\citep{bao2023fast, sadasivan2023can, LIANG2023100779,wang-etal-2024-m4}.

Wikipedia is a longstanding, publicly-curated reference source for an expansive and ever-growing set of topics. In the era of LLMs, it has become a standard source of training data due to its breadth of information, standards of curation, and flexible licensing. 
Therefore, it is an important testing ground for the proliferation of AI-generated content. 
We collect Wikipedia pages created in August 2024 and use a previously curated dataset of pages created prior to March 2022 as a pre-GPT-3.5 baseline for our experiments (\autoref{sec:sources}).\footnote{Our data collection and evaluation code is available at \url{github.com/brooksca3/wiki_collection}.}
We detect a noticeable increase in AI-generated content in the 2024 data and qualitatively assess flagged articles
(\autoref{sec:analysis}). We compare these findings with preliminary experiments conducted on other contemporary sources (\autoref{sec:other}) and comment on the implications of AI-generated content (\autoref{sec:implications}).


\section{Detection Tools}
\label{sec:tools}

We use two prominent detection tools which were suitably scalable for our study. \texttt{GPTZero}~\citep{tian2023gptzero} is a commercial AI detector that reports the probabilities that an input text is entirely written by AI, entirely written by humans, or written by a combination of AI and humans. In our experiments we use the probability that an input text is entirely written by AI. The black-box nature of the tool limits any insight into its methodology.

An open-source method, \texttt{Binoculars}~\citep{hans2024spotting} uses two separate LLMs $\mathcal{M}_1$ $\mathcal{M}_2$ to score a text $s$ for AI-likelihood by normalizing perplexity by a quantity termed cross-perplexity, which computes the average cross-entropy between the outputs of two models over a span of tokens: \\
\[
B_{\mathcal{M}_1, \mathcal{M}_2}(s) = \frac{\log \text{PPL}_{\mathcal{M}_1}(s)}{\log \text{X-PPL}_{\mathcal{M}_1, \mathcal{M}_2}(s)}
\]
 The input text is classified as AI-generated if the score is lower than a determined threshold, calibrated according to a desired false positive rate (FPR). 
For our experiments, we use Falcon-7b and Falcon-7b-instruct \citep{almazrouei2023falcon} to calculate cross-perplexity, following \citet{hans2024spotting} who report it as the best pair of LLMs for detection.
Compared to competing open-source detectors, \texttt{Binoculars} reports superior performance across various domains including Wikipedia \citep{hans2024spotting}.

\section{Wikipedia Data Sources}
\label{sec:sources}

Wikipedia provides an accessible list of articles created within the past month for supported languages.
We use the \textit{New Pages} feature to collect articles created in August 2024 in English, French, German, and Italian (Table~\ref{tab:wiki_stats}). These languages were also available in a set of Wikipedia pages collected before March 2022.\footnote{\url{https://huggingface.co/datasets/legacy-datasets/wikipedia}}

Although GPT-3 was released in June 2020, the significant public uptake in generating text with LLMs occurred in March 2022 with the release of GPT-3.5 and exploded with ChatGPT in November 2022 \citep{wu2023brief}. Thus, the dataset of articles created prior to March 2022 allows us to establish a FPR for the tools in detecting AI-generated content post-GPT-3.5. 

\begin{table}[h!]

    \small
    \centering
    \begin{tabular}{lcc}
        \toprule
        \textbf{Language} & \textbf{Pre-March 2022} & \textbf{August 2024} \\
        \midrule
        English & 2965 & 2909 \\
        German  & 4399 & 3907 \\
        Italian & 2306 & 3003 \\
        French  & 4351 & 3138 \\
        \bottomrule
    \end{tabular}
    \caption{Number of Wikipedia pages collected for each language before March 2022 and in August 2024 after removing articles containing fewer than $100$ words. We take random subsets of our data pools to stay within budget constraints.}
    \label{tab:page_creation}
\end{table}

\begin{table*}[]
    \small
    \centering
    \begin{tabular}{lcccc}
        \toprule
        \textbf{Language} & \multicolumn{2}{c}{\textbf{Footnotes per Sentence}} & \multicolumn{2}{c}{\textbf{Outgoing Links per Word}} \\
        \cmidrule(lr){2-3} \cmidrule(lr){4-5}
        & \textbf{AI-Detected Articles} & \textbf{All New Articles} & \textbf{AI-Detected Articles} & \textbf{All New Articles} \\
        \midrule
        English & 0.667 & \textbf{0.972} & 0.383 & \textbf{1.77} \\
        French  & 0.370 & \textbf{0.441} & 0.474 & \textbf{1.58} \\
        German  & 0.180 & \textbf{0.211} & 0.382 & \textbf{0.754} \\
        Italian & \textbf{0.549} & 0.501 & 1.16 & \textbf{1.64} \\
        \bottomrule
    \end{tabular}
    \caption{Mean values for footnotes per sentence and outgoing links per word in all articles created in August 2024, as well as those detected as AI-generated by either \texttt{GPTZero} or \texttt{Binoculars}, with thresholds set to induce a 1\% FPR for each tool. The number of AI articles are $207$, $174$, $249$, and $206$ for English, French, German, and Italian.}
    \label{tab:wiki_stats}
\end{table*}

\section{Detection as a Lower Bound}
Following \citeposs{Latona2024TheAR} approach for measuring AI content in conference reviews, we estimate a lower bound for AI-generated articles by subtracting the percentage of pre-March 2022 articles classified as AI by a given tool from the percentage of August 2024 articles classified as AI. As we do not have ground-truth examples of AI-generated articles, we do not attempt to estimate the false negative rate (FNR). Doing so would require creating artificial positive examples by simulating the various ways Wikipedia authors might use LLMs to assist in writing---taking into account different models, prompts, and the extent of human integration, among other factors.

Although we cannot speculate on how \texttt{GPTZero} scores text, Falcon models are trained on Wikipedia data \citep{almazrouei2023falcon}, and \texttt{Binoculars} is known to assign false positives to text in its models' training data \citep{hans2024spotting}. Additionally, the tools we use are primarily for detecting AI-generated content in English. While \texttt{GPTZero} supports Spanish and French, it is not designed for other languages \citep{GPTZero_Multilingual2024}, and using it out-of-domain may increase FNRs. For non-English texts, \texttt{Binoculars} reports similar FPRs but higher FNRs \citep{hans2024spotting}. The higher the FNRs, the more AI-generated articles slip past the detectors. Therefore, while the numbers we report represent a lower bound, the actual amount of AI-generated content could be substantially higher.

Our methodology assumes that the pre-March 2022 and August 2024 data distributions are comparable, with increased AI use being the primary factor affecting detection. One concern is that pre-March 2022 pages may be more polished due to years of editing. However, we observe that a higher number of edits weakly correlates with a higher AI-detection score for pre-March 2022 articles (\autoref{sec:edits}), suggesting that the FPRs for those articles may even be inflated.  While the base assumption cannot be watertight, we observe a relatively consistent distribution of page categories between the two data pools, and we rely on the consistency of our chosen tools' reported FPRs.

%

%

\section{Trends in Pages Flagged for AI}
\label{sec:analysis}

 As seen in \autoref{fig:wiki}, we estimate that 4.36\% of 2,909 English Wikipedia articles created in August 2024 contain significant AI-generated content.\footnote{5.36\% detection rate with 1\% FPR.} We set the classification thresholds of both tools to induce a detection rate of no more than 1\% on pre-March 2022 articles. With these thresholds, \texttt{GPTZero} classifies $156$ English articles as AI-generated, and \texttt{Binoculars} classifies $96$. Among these, there is an overlap of $45$ articles classified as AI independently by the two tools. Notably, there is no overlap between the $39$ and $31$ pre-March 2022 English articles flagged as AI-generated by the tools. Hence, there is a strong shared signal in assumed true positives but tool-specific noise in false positives. 

The quality of articles detected as AI-generated is generally lower on at least two axes. \autoref{tab:wiki_stats} shows how, compared to all articles created in August 2024, AI-generated ones use fewer references and are less integrated into the Wikipedia nexus.\footnote{We normalize by sentence and word count to remove length as a confounding factor, since longer articles may have more footnotes and links without being higher quality.}

\subsection{Manual Inspection}

\begin{figure*}[t!]
    \centering
    \includegraphics[width=\linewidth]{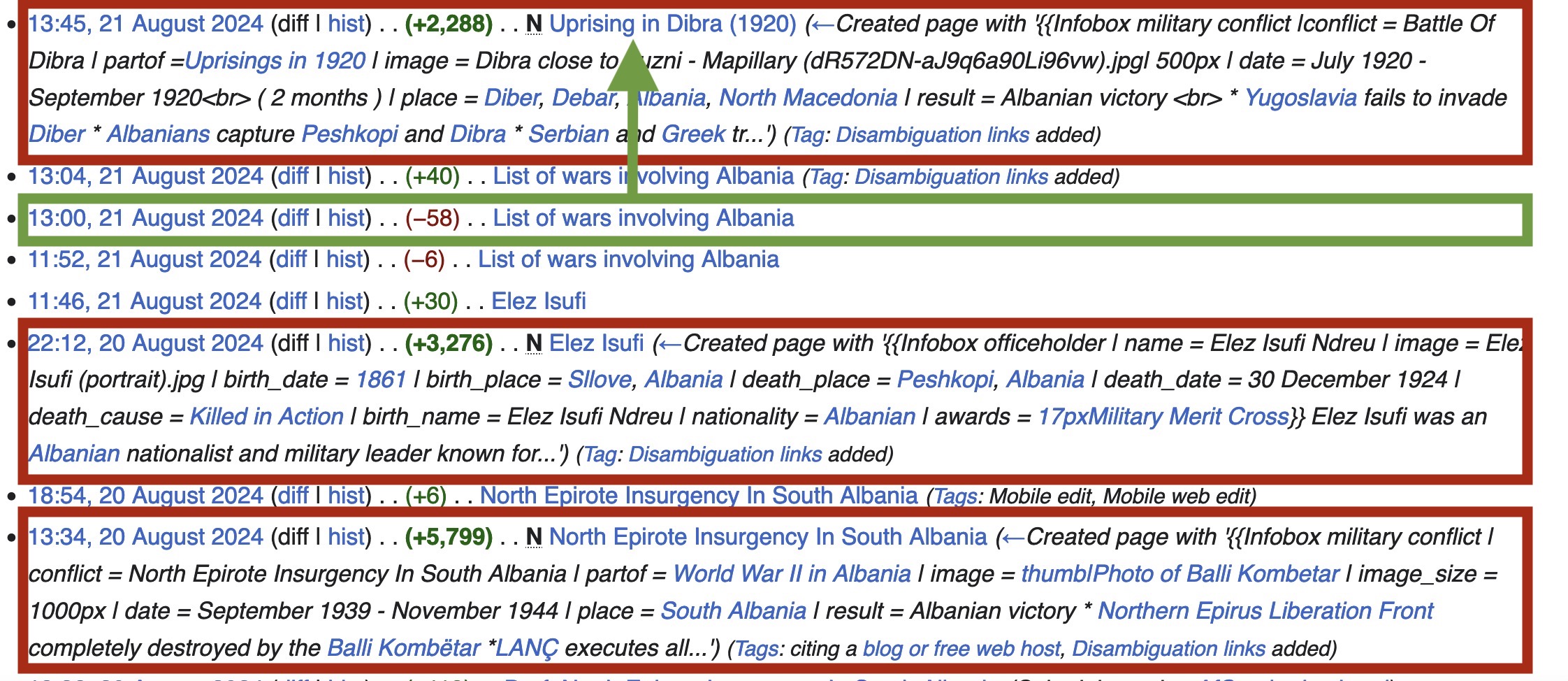}
    \caption{The activity of this user, who was flagged for instigating an `Edit War,' reveals that within a single day, they created three articles (red border), all identified as AI-generated. Notably, at 13:00 (green border), the user edited the outcome of `War in Dibra' from \textit{`Mixed Results'} to \textit{`Victory'} and removed key text, just an hour before creating a new page titled `Uprising in Dibra.' That page (see \autoref{fig:albania}) has since been deleted by moderators.}
    \label{fig:alb_edits}
\end{figure*}

We inspect each of the $45$ English articles flagged as AI-generated by both \texttt{GPTZero} and \texttt{Binoculars} by examining their edit histories and the activity logs of their creators to better understand the motivations for using LLMs to create Wikipedia pages. 
%
%
We observe that several of the $45$ pages are authored by the same individuals, which is unsurprising, as users who use AI in one article are likely to use it in others. Most of the $45$ pages are flagged by moderators and bots with some warning, e.g., ``\textit{This article does not cite any sources. Please help improve this article by adding citations to reliable sources}'' or even ``\textit{This article may incorporate text from a large language model.}'' We observe distinct trends after inspecting the user and page histories.

\subsection{Advertisement}
One prominent motive is self-promotion. Of the $45$ flagged pages, we identify eight that were created to promote organizations such as small businesses, restaurants, or websites. These pages are often the first to be created by their respective users and typically lack any citations beyond links to the entity being promoted. One page links to a personal YouTube video promoting a winery with fewer than $100$ views. Another describes an estate in the United Kingdom, claiming it has formerly had notable residents.  This is subsequently deleted by a moderator who notes:
\begin{quote} 
``\textit{Reference links are all dead apart from one for the town council, which makes no mention of the estate. One link is actually labelled `fictional'... Article reads like an advert for the house, which is coincidentally up for sale at the moment.}''
\end{quote} 
Other self-promoting pages are deleted by moderators who remark: ``\textit{unambiguous advertising which only promotes a company, group, product, service, person, or point of view.}''

\subsection{Pages Pushing Polarization}
While the immediate beneficiaries of advertisement are obvious, we also identify pages that advocate a particular viewpoint on often polarizing political topics.
We identify eight such pages out of the flagged $45$. 
One user created five articles on English Wikipedia, detected by both tools as AI-generated, on contentious moments in Albanian history. The same user's profile garnered a warning from Wikipedia for engaging in an `Edit War' with other users  (\autoref{fig:alb_edits}). The user changed outcomes of an Albanian conflict from \textit{`Mixed Results'} to \textit{`Victory'} and deleted supporting text, before using AI to generate an entirely new page on said conflict. The Wikimedia community has since removed the flagged pages and banned the user in question for sockpuppetry.\footnote{Sockpuppetry is the practice of using multiple accounts to mislead other editors~\citep{solorio2013case}.} In other cases, users created articles ostensibly on one topic, such as types of weapons or political movements, but dedicated the majority of the pages' content to discussing specific political figures. We find two such articles that espouse non-neutral views on JD Vance and Volodymyr Zelensky.

\subsection{Machine Translation}
AI detection tools can flag instances of machine translation. We find three cases where users explicitly documented their work as translations, including pages on Portuguese history and legal cases in Ghana. Outside of the $45$ articles flagged by both tools, we identify a top contributor of Italian Wikipedia who created $57$ articles flagged as AI-generated by \texttt{Binoculars}, but not by \texttt{GPTZero}.\footnote{These $57$ translated pages are the reason \texttt{Binoculars} has a higher detection rate than \texttt{GPTZero} for Italian in \autoref{fig:wiki}.} This user notes in their sandbox that they translated these articles from French Wikipedia, a common practice in the Wikimedia editor community \citep{wikimedia}. 

Despite producing fluent and accurate translations, state-of-the-art LLMs still introduce observable biases \citep{hendy2023good}. Even beyond these biases, machine translation complicates the process of vetting pages flagged for AI content: an AI-generated article in one language can be translated and propagated into other languages. For example, Wikipedia communities like Cebuano and Swedish contain millions of pages made through automatic methods~\citep{alshahrani2023depth}.


\subsection{Writing Tool} 
Other pages, which are often well-structured with high-quality citations, seem to have been written by users who are knowledgeable in certain niches and are employing an LLM as a writing tool. Several of the flagged pages are created by users who churn out dozens of articles within specific categories, including snake breeds, types of fungi, Indian cuisine, and American football players. One flagged page points us to a user who seemingly uses AI to create chapter-by-chapter books summaries. Another page details an ongoing criminal case in India and is flagged by moderators with a warning reminding editors to treat subjects as innocent until proven guilty.

\section{Detection Beyond Wikipedia}
\label{sec:other}
%
%
%
Wikipedia has a distinct genre and brand of contributor. To contextualize our findings and motivate further research, we conduct a preliminary investigation into two other genres---comment-section debates and press releases---on platforms where contributors may have different motivations for using generative AI compared to those on Wikipedia. We hope this encourages closer examination of AI-generated content across different domains with varying contributor incentives.\footnote{Full details about the sources we evaluated and instructions for replicating the evaluation are available at our repository:  \url{github.com/brooksca3/wiki_collection}.}

\subsection{Reddit}
 Comments on contentious subreddits---Israel-Palestine, public opinion on Democrats, public opinion on Republicans---are updated daily on Kaggle, a popular data science platform. We randomly sample 3,000 user comments from 2024 containing at least 100 words.

Less than $1\%$ of the collected comments receive a \texttt{GPTZero} score above $0.5$, which may mean $(1)$ few are AI-generated, $(2)$ such content is censored or $(3)$ AI presence is difficult for detectors to discern in this domain. Despite being rare, some comments flagged as AI-generated are potentially worrisome: one urges others how to vote in an upcoming election~(\autoref{sec:reddit}).

\subsection{Press Releases}
The United Nations "remains the one place on Earth where all the world’s nations can gather together, discuss common problems, and find shared solutions that benefit all of humanity".\footnote{\url{https://www.un.org/}}  Country teams of the United Nations provide frequent updates about developments in that country.  We collect 8,326 press releases across 60 country teams from the United Nations from 2013 to 2024; country teams have websites in the format of https://\{country\}.un.org.\footnote{Due to licensing uncertainties, we do not release the press releases; however, we release the scripts used to collect them.}

As many as 20\% of press releases published in 2024 received a \texttt{GPTZero} AI-generation score of at least 0.5, compared to 12.5\% in 2023, 1.6\% in 2022, and less than 1\% in all years prior.\footnote{$90/447$ press releases from 2024 are flagged, $170/1360$ from 2023, and $20/1268$ from 2022.} The marked increase in UN press releases flagged as AI may stem from translations into English, although the individuals named as authors of the articles often hold degrees from institutions in English-speaking countries.
We include three examples of flagged press releases in \autoref{sec:un_samples}.

\section{Implications and Conclusion}
\label{sec:implications}

Not all AI-generated text is nefarious. If a human authors the primary content and approves an AI-generated summary or translation, AI may be considered a writing aide. \citet{shao2024assisting} have even designed a retrieval-based LLM workflow for writing Wikipedia-like articles and gathered perspectives from experienced Wikipedia editors on using it---the editors unanimously agreed that it would be helpful in their pre-writing stage. Moreover, LLM-enabled translation can reduce language barriers in domains of information sharing \citep{katsnelson2022poor, berdejo2023ai}.

However, the increasing ease with which it is possible to generate content at scale to overrepresent a particular perspective has predictable and dangerous consequences. People are more likely to believe statements that are frequently repeated, since familiarity is easily confused with validity \citep{hasher1977frequency,unkelbach2019truth}. Consumer confidence is a key determiner of economic strength, and confidence in the economy is based in part on how strong individuals perceive others’ confidence to be. To the extent that AI-generated outputs show less variability than human-generated texts, we can expect peaks of polarization to continue to increase \cite{bail2018exposure,heltzel2020polarization}, undermining the useful wisdom of crowds \citep{surowiecki2005wisdom,bender2021dangers}.

Continued work is needed to understand differences in LLMs and human speech and the implications of widespread AI-generated data~\citep{guo-etal-2023-hc3,sadasivan2023can,liang2024mapping}. The motives to discreetly propagate AI-generated text online vary across platforms, and measuring the prevalence of AI-generated content is a necessary step in understanding these motives. 

\newpage
\section*{Limitations}
The proprietary nature of \texttt{GPTZero} makes experiments costly to run (\$1000 for our study).
\texttt{Binoculars} requires non-trivial RAM and compute to run at scale. These factors bound the scale of the study we are able to conduct and limit our ability to draw generalizable conclusions. We hope that future efforts can replicate this work at a larger scale and across more domains. 

Future work should also consider a broader suite of AI detectors. We considered two other open-source AI detection tools but did not use them. \texttt{Ghostbuster} \citep{verma2023ghostbuster} requires training on specific LLM features and \texttt{Fast-DetectGPT} \citep{bao2023fast} reports lower true positive rates than  \texttt{Binoculars} across all domains considered.

Moreover, we focus on English and other high resource languages given their availability in the sources we consider. In the multilingual setting,
\citet{LIANG2023100779} detect bias in AI detectors against non-native speakers, \citet{wang-etal-2024-m4} create a multilingual dataset to study detection, and \citet{ignat2024maide} study multilingual detection in the context of hotel reviews. 


\section*{Ethical Considerations}
Detecting AI may have unexpected negative consequences for people implicated as having generated that text. We have therefore been encouraged to omit any identifying information in the specific pages we discuss; however, we will provide more specific data to researchers upon request provided that it not be disseminated further.

We are relying on public internet content.  All sources that we investigate are public-facing in nature. 
The Wikipedia data we collect is under a Creative Commons CC0 License. 
The Reddit data is distributed through Kaggle under a Open Data Commons Attribution License (ODC-By) v1.0.  
There is no clear license for United Nations country teams.  Individual use \textit{and} download of the data is explicitly permitted by the parent organization.

\section*{Acknowledgements}
We thank Adele Goldberg for funding support along with Brandon Stewart, Preeti Chemiti, and Rob Voigt for valuable feedback and advice. We are grateful to the Princeton Language and Intelligence (PLI) for providing computational resources. Peskoff is supported by the National Science Foundation under Grant
\#2127309 to the Computing Research Association
for the CIFellows 2021 Project.

\bibliography{acl_latex}
\clearpage
\onecolumn
\begin{center}
\Large \textbf{Appendix}
\end{center}
\appendix

\section{(Deleted) Wikipedia Page Classified as AI-Generated}
\begin{figure*}[h]
  \centering
  \includegraphics[width=.9\textwidth]{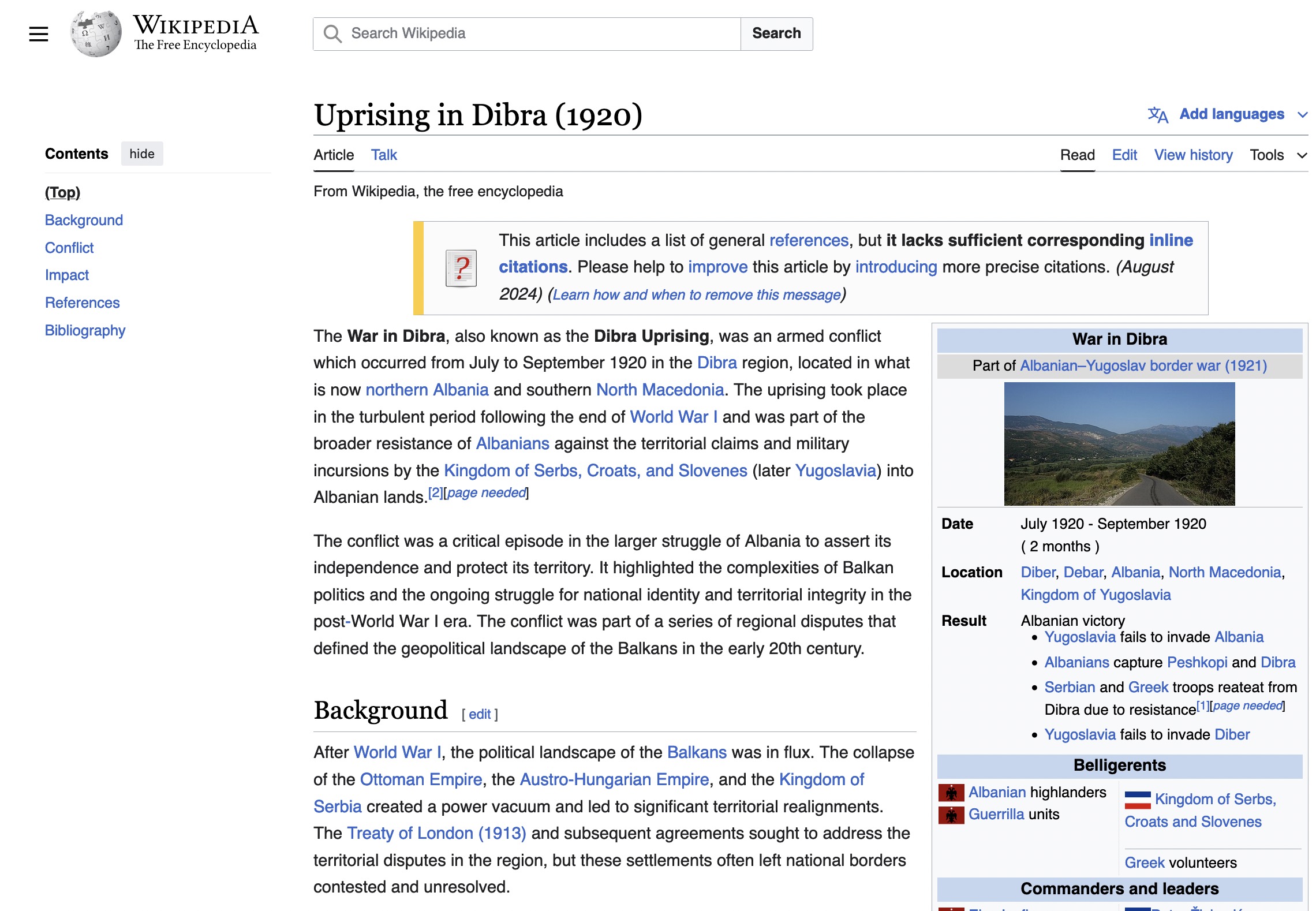} 
  \caption{Wikipedia page flagged as AI-generated and deleted by moderators.}
  \label{fig:albania}
\end{figure*}

\section{Reddit Post Classified as AI-Generated}
\label{sec:reddit}

The following comment encouraging Americans to vote for a third-party candidate was flagged as AI.
\begin{quote}
\textit{While the acknowledgment of the symbolic rejection of the two-party system is understood, the contention here lies in the practical consequences of a third-party vote. It's crucial to recognize that the call for voting third party isn't solely symbolic but a strategic push for a more diverse political landscape over time.  
The argument asserts that voting for anyone other than Biden increases Trump's chance of victory. However, this perspective assumes a binary outcome, overlooking the potential long-term impact of promoting alternative voices. A shift toward a multi-party system is a gradual process, and fostering this change requires voters to make choices aligned with their principles.
Moreover, characterizing the choice between a ``bland moderate Democrat" and an ``extremely corrupt, authoritarian Republican" as high stakes underscores the need for broader political options. Supporting third parties now can pave the way for a more representative democracy in the future, where voters aren't limited to perceived lesser evils.
While the current election might seem high-stakes, it's crucial to consider the long-term goal of breaking the duopoly for a healthier democracy. Third-party votes, rather than being mere protests, can be strategic steps toward that transformative change.}
\end{quote}

\clearpage

\clearpage
\section{Examples of UN Press Releases Classified as AI-Generated}
\label{sec:un_samples}
In this section, we present three examples of UN press releases flagged by our tools as likely AI-generated. We re-emphasize that AI detection can produce false positives, and no individual classification should be considered definitive.

\subsection{UN Belize Press Release} \label{sec:belize}

\begin{table}[htbp]
\captionsetup{width=0.75\textwidth}
\captionsetup{justification=centering}
\centering
\begin{tabularx}{\textwidth}{X}
\hline \\
\large
\textit{The United Nations in Belize expresses its deep concern over the recent tragic incidents that have claimed the lives of women and children both in their homes and public spaces}
\\ \\
\url{https://belize.un.org/en/263463-united-nations-belize-expresses-its-deep-concern-over-recent-tragic-incidents-have-claimed}

\\ \\

\texttt{The United Nations in Belize expresses its deep concern over the recent tragic incidents that have claimed the lives of women and children both in their homes and public spaces. The right to life is fundamental and should be expected and respected by all in Belize. We offer our condolences to families affected by these recent tragic cases of domestic and gender-based violence and commit to continue supporting the Government and people of Belize in the pursuit of freedom from violence. We all collectively have a role to play in ensuring that Belize remains a safe, secure, and inclusive society for everyone. The United Nations works to support Belize's commitment to eliminate all forms of violence especially against women and girls making the recent events even more distressing. The United Nations is fully committed to support the Government of Belize and civil society in concrete actions to realize the rights of all women and children, allowing them to live lives free of violence including preventive support and the attention of mental health aspects and consequences of those affected.} 
\caption{Press Release by the United Nations in Belize, 15 March 2024}
\label{fig:belize}
\end{tabularx}
\end{table}

\clearpage

\subsection{(Abridged) UN Bangladesh Press}
\label{sec:bang_press}
\begin{table}[htbp]
\captionsetup{width=0.75\textwidth}
\captionsetup{justification=centering}
\centering
\label{tab:full_prompt_example_alb}
\begin{tabularx}{\textwidth}{X}
\hline \\
\large
\textit{UNOPS' Roundtable Discussion on the `Invest in  Women: Accelerate Progress'}
\\ \\
\url{https://bangladesh.un.org/en/264789-unops-roundtable-discussion-\%C2\%A0\%E2\%80\%98invest-\%C2\%A0women-accelerate-progress\%E2\%80\%99}
\\ \\ 

\texttt{Dhaka, Bangladesh - UNOPS Bangladesh hosted the 9th episode of "SDG Café," a monthly roundtable discussion series dedicated to addressing pressing development challenges and co-creating innovative solutions. \newline \newline As part of UNOPS’s commitment to getting Agenda 2030 back on track, this episode places the spotlight on the Sustainable Development Goals (SDG 5), dedicated to advancing gender equality and empowering women in Bangladesh and beyond. This roundtable took place on March 21, 2024 with the theme, ‘Invest in Women: Accelerate Progress’. \newline \newline The session focused on highlighting the importance of investing in women to foster inclusive and sustainable economic growth, in line with SDG 5. Addressing the enduring gender disparities in investment, especially in developing nations, the talks revolved around discussing obstacles, prospects, and inventive approaches to boost investment in businesses owned by women, elevate women into leadership positions, and advance initiatives supporting gender parity. \newline \newline The highlight of the event was the keynote speeches delivered by esteemed personalities Rubana Huq, Vice-chancellor of Asian University for Women and Chairperson of Mohammadi Group, and Azmeri Haque Badhon, renowned Bangladeshi actress. Huq's address emphasized the urgency of accelerating investment in women, drawing from her extensive experience in academia and business leadership.} \newline $\hdots$\\ \hline 

\caption{Press Release by the United Nations in Bangladesh, 2 May 2024}
\end{tabularx}
\end{table}

\clearpage

\subsection{(Abridged) UN Turkmenistan Press Release}
\label{sec:turk_press}
\begin{table}[htbp]
\captionsetup{width=0.75\textwidth}
\captionsetup{justification=centering}
\centering
\begin{tabularx}{\textwidth}{X}
\hline \\
\large
\textit{Consultative meeting with national stakeholders on Advancing Cross-Border Paperless Trade in Turkmenistan}
\\ \\
\url{https://turkmenistan.un.org/en/269295-consultative-meeting-national-stakeholders-advancing-cross-border-paperless-trade}
\\ \\ 

\texttt{Turkmenistan, Ashgabat - The United Nations Resident Coordinator's Office (UN RCO) in Turkmenistan and the United Nations Economic and Social Commission for Asia and the Pacific (ESCAP) jointly organized a two-day workshop titled "Towards a National Strategy in Advancing Cross-Border Paperless Trade in Turkmenistan." The event, held on 20-21 May 2024 at the UN House in Ashgabat, brought together national stakeholders and development partners to discuss and strategize the implementation of cross-border paperless trade initiatives in Turkmenistan.
The opening day of the workshop featured esteemed speakers including Ms. Rupa Chanda, Director of Trade, Investment and Innovation Division at ESCAP, Mr. Dmitry Shlapachenko, UN Resident Coordinator in Turkmenistan, and Mr. Myrat Myradov, Head of Legal Regulations and Coordination at the Foreign Economic Relations Department, Ministry of Trade and Foreign Economic Relations of Turkmenistan. \newline\newline The first day's sessions included a comprehensive review of key initiatives by various ministries and agencies, aimed at enhancing trade facilitation in Turkmenistan. Development partners, Asian Development Bank, USAID, GIZ, International Trade Center, also presented their contributions in this domain, fostering a better understanding of the current trade facilitation landscape in the country... \newline\newline The workshop concluded with a practical group exercise, followed by group presentations, and summarizing the outcomes and proposed strategies for advancing cross-border paperless trade in Turkmenistan.
The event underscored Turkmenistan's commitment to embracing innovative solutions for trade facilitation and integration into the global digital economy.
Turkmenistan joined the CPTA in May 2022 and has actively participated in its implementation. A readiness assessment was conducted, resulting in a study report published in December 2022.} \newline \\ \hline 

\caption{Press Release by the United Nations in Turkmenistan, 22 May 2024}
\end{tabularx}
\end{table}

\clearpage

\section{AI Detection Scores vs. Page Edits Across Languages} \label{sec:edits}

\begin{figure*}[h]
  \centering
  \includegraphics[width=.48\textwidth]{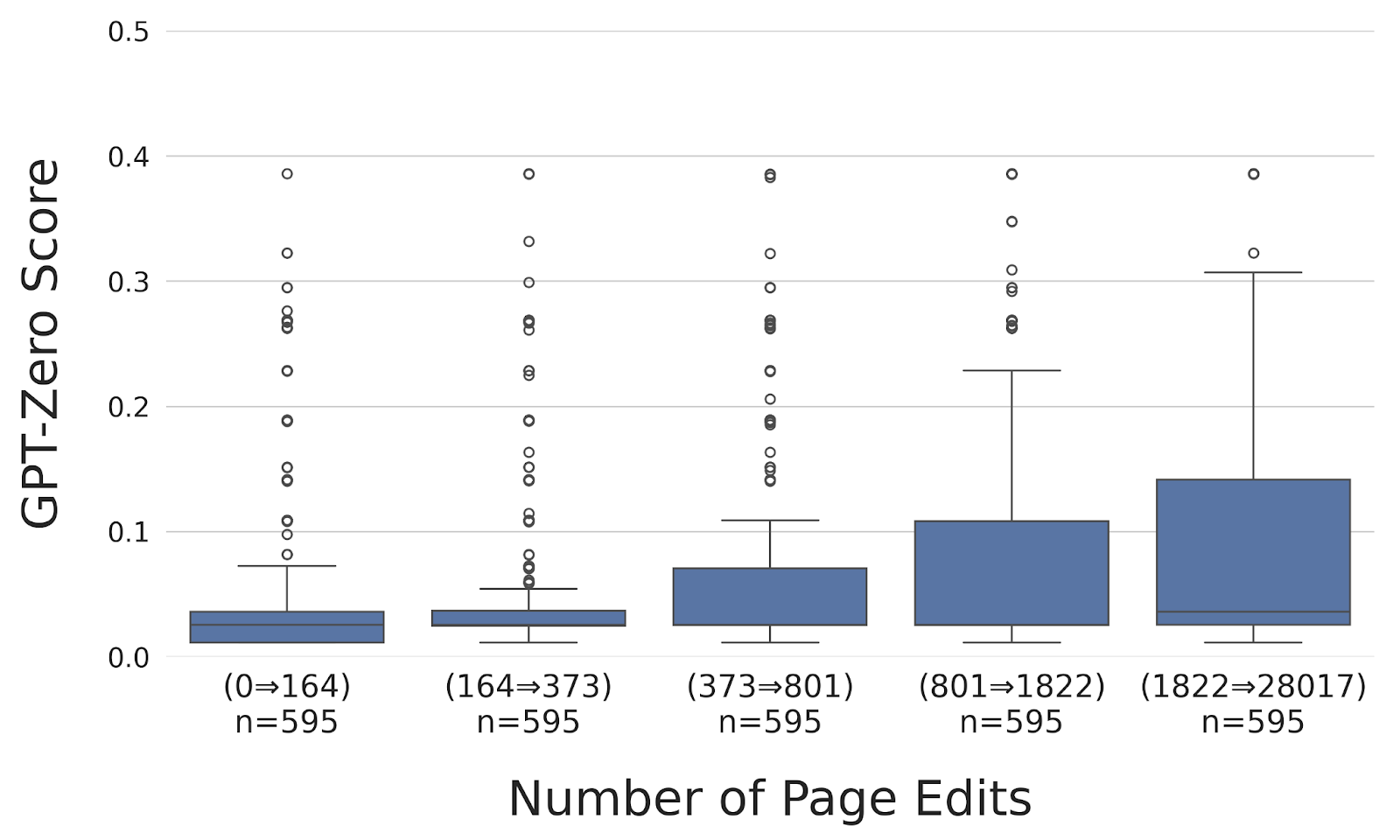}
  \hfill
  \includegraphics[width=.48\textwidth]{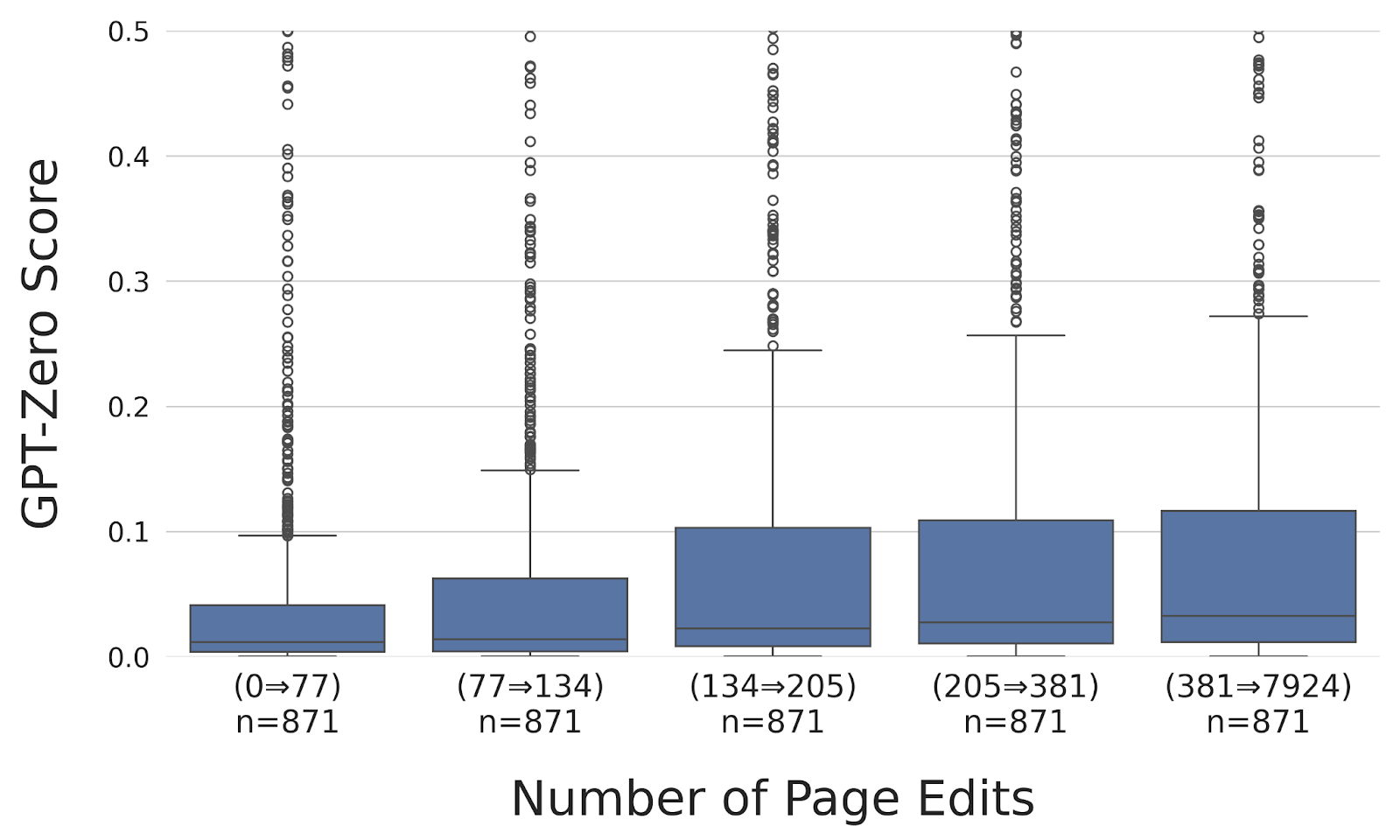}
  \caption{\texttt{GPTZero} scores compared to the number of page edits for English (left) and French (right) articles created before March 2022.  Pages with more edits in English receive higher \texttt{GPTZero} scores.}
  \label{fig:en_fr_edits}
\end{figure*}

\begin{figure*}[h]
  \centering
  \includegraphics[width=.48\textwidth]{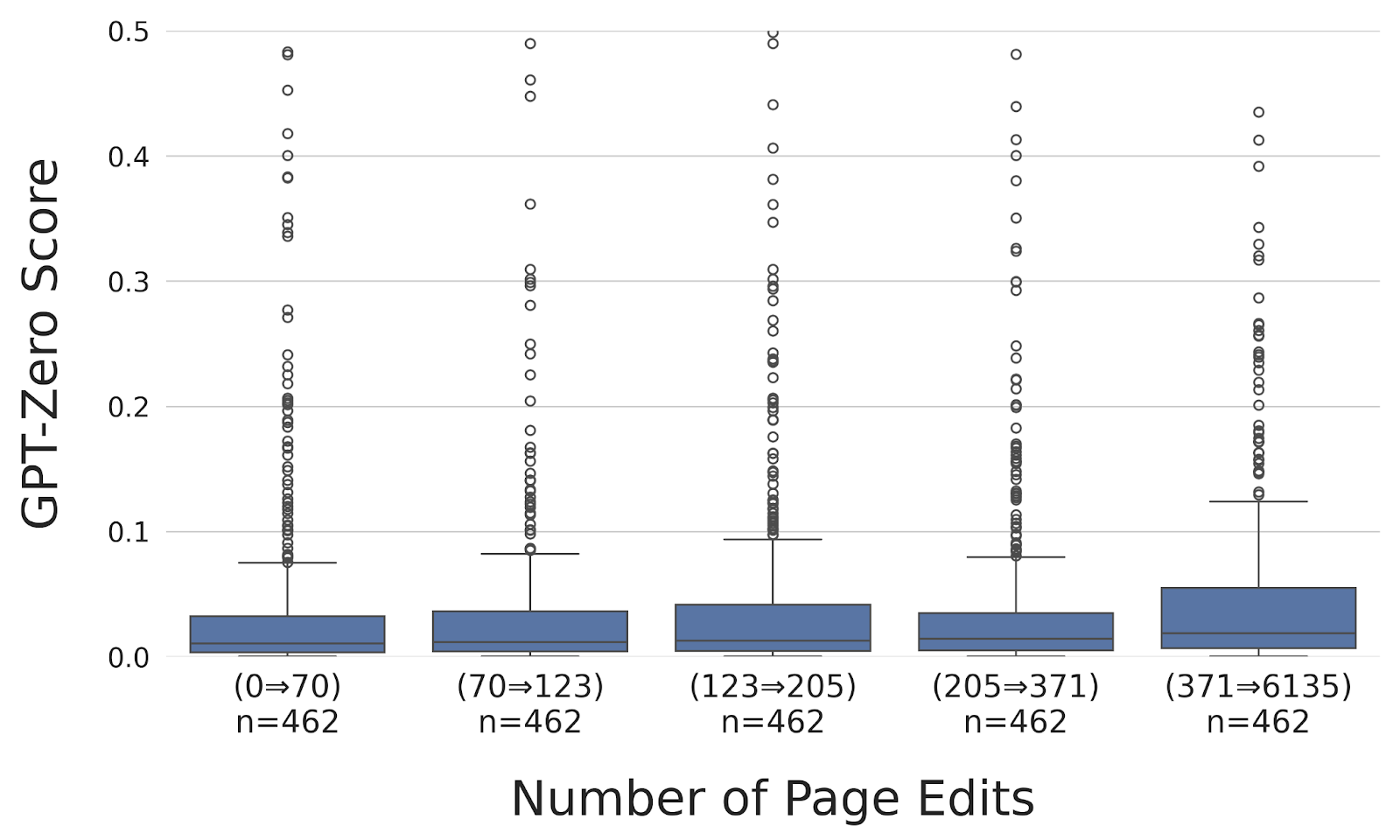}
  \hfill
  \includegraphics[width=.48\textwidth]{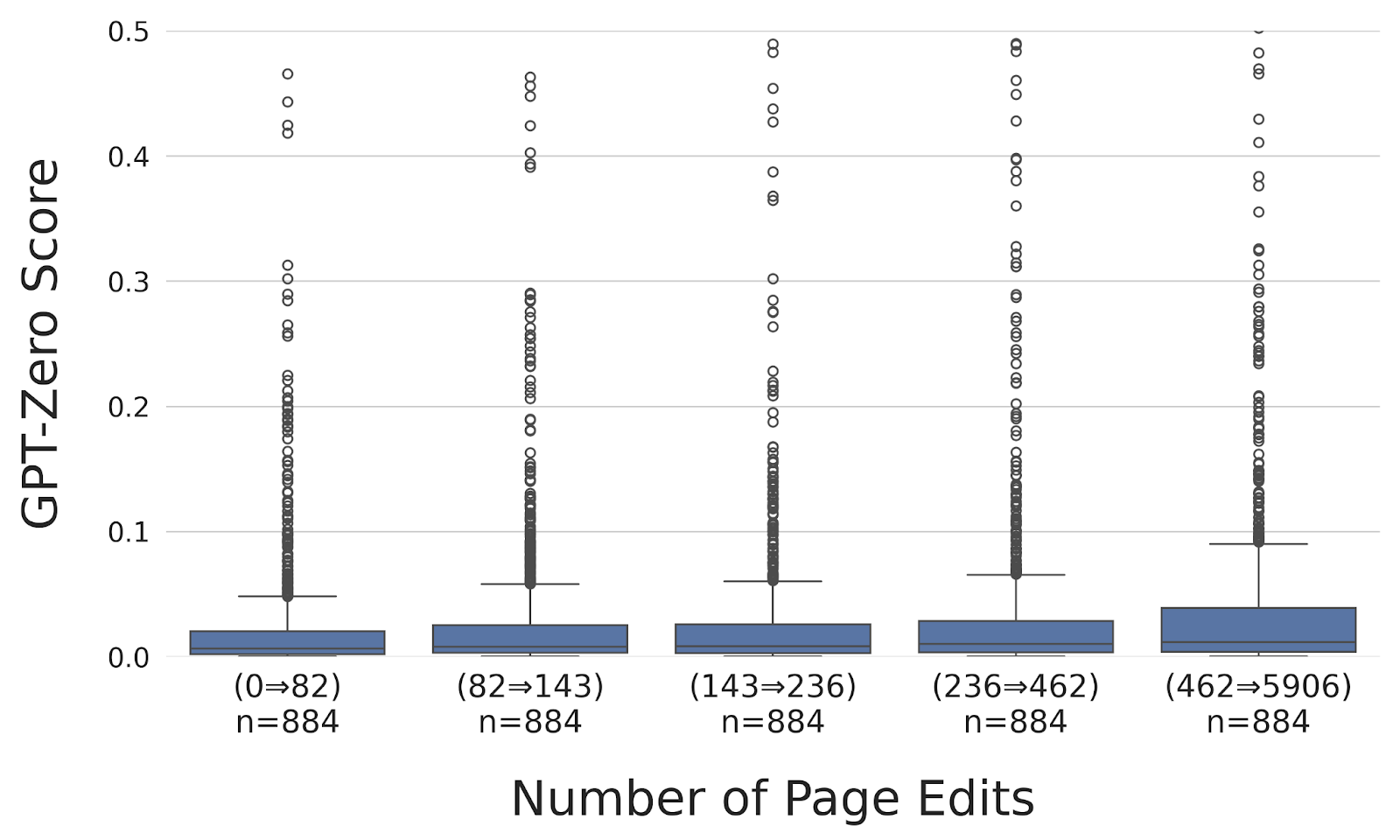}
  \caption{\texttt{GPTZero} scores compared to the number of page edits for Italian (left) and German (right) articles created before March 2022.}
  \label{fig:it_de_edits}
\end{figure*}

\end{document}